\documentclass[10pt,twocolumn,letterpaper]{article}
\usepackage{xcolor}
\usepackage[accsupp]{axessibility}  

\usepackage{cvpr}              

\usepackage{booktabs}
\usepackage{multirow}
\usepackage{array}
\usepackage{colortbl}
\usepackage{tabularx}
\usepackage{pifont}
\DeclareMathOperator*{\argmax}{arg\,max}
\DeclareMathOperator*{\argmin}{arg\,min}

\definecolor{cvprblue}{rgb}{0.21,0.49,0.74}
\usepackage[pagebackref,breaklinks,colorlinks,allcolors=cvprblue]{hyperref}

\def\paperID{14486} 
\def\confName{CVPR}
\def\confYear{2026}

\title{ConceptPose: Training-Free Zero-Shot Object Pose Estimation \\ using Concept Vectors}

\author{
Liming Kuang\textsuperscript{1,2} \,
Yordanka Velikova\textsuperscript{1,2} \,
Mahdi Saleh\textsuperscript{1} \,
Jan-Nico Zaech\textsuperscript{3} \, \\
Danda Pani Paudel\textsuperscript{3} \,
Benjamin Busam\textsuperscript{1,2}
\\
\textsuperscript{1}Technical University of Munich \quad
\textsuperscript{2}Munich Center for Machine Learning
\\
\textsuperscript{3}INSAIT, Sofia University ``St. Kliment Ohridski''
\\
{\tt\small \{liming.kuang, yordanka.velikova, m.saleh, b.busam\}@tum.de}
\\
{\tt\small \{jan-nico.zaech, danda.paudel\}@insait.ai}
}

\begin{document}
\maketitle

\let\thefootnote\relax\footnotetext{Project page: \url{https://stevenlk.xyz/conceptpose}.}

\begin{abstract}
Object pose estimation is a fundamental task in computer vision and robotics, yet most methods require extensive, dataset-specific training. Concurrently, large-scale vision language models show remarkable zero-shot capabilities.
In this work, we bridge these two worlds by introducing ConceptPose, a framework for object pose estimation that is both training-free and model-free. ConceptPose leverages a vision-language-model (VLM) to create open-vocabulary 3D concept maps, where each point is tagged with a concept vector derived from saliency maps. By establishing robust 3D-3D correspondences across concept maps, our approach allows precise estimation of 6DoF relative pose. Without any object or dataset-specific training, our approach achieves state-of-the-art results on common zero shot relative pose estimation benchmarks, outperforming the strongest baseline by a relative 62\% in average ADD(-S) score, including methods that utilize extensive dataset-specific training.
\end{abstract}    
\section{Introduction}
\label{sec:intro}
Making machines see and think in 3D requires precise object understanding. In the era of embodied AI, object pose estimation has therefore become a critical building block for agents interacting with the physical world. The task is to determine an object's 6D pose from an image. It enables key capabilities across robotic manipulation \cite{wen2021catgrasp, wang2019densefusion, 6dpose_robotic_survey}, augmented reality \cite{ar_6dpose, Su2019DeepMO}, and autonomous navigation \cite{autonom_driving_6dpose}. However, classical approaches have long been constrained by a critical bottleneck: they require extensive, object-specific training, often depending on precise 3D CAD models and large datasets of ground-truth poses~\cite{hinterstoisser2013,xiang2018,wang2019,hodan2018}. This reliance on pre-trained models and curated data fundamentally limits their ability to adapt to novel objects and environments.

\begin{figure}[t]
    \centering
    \includegraphics[width=\linewidth]{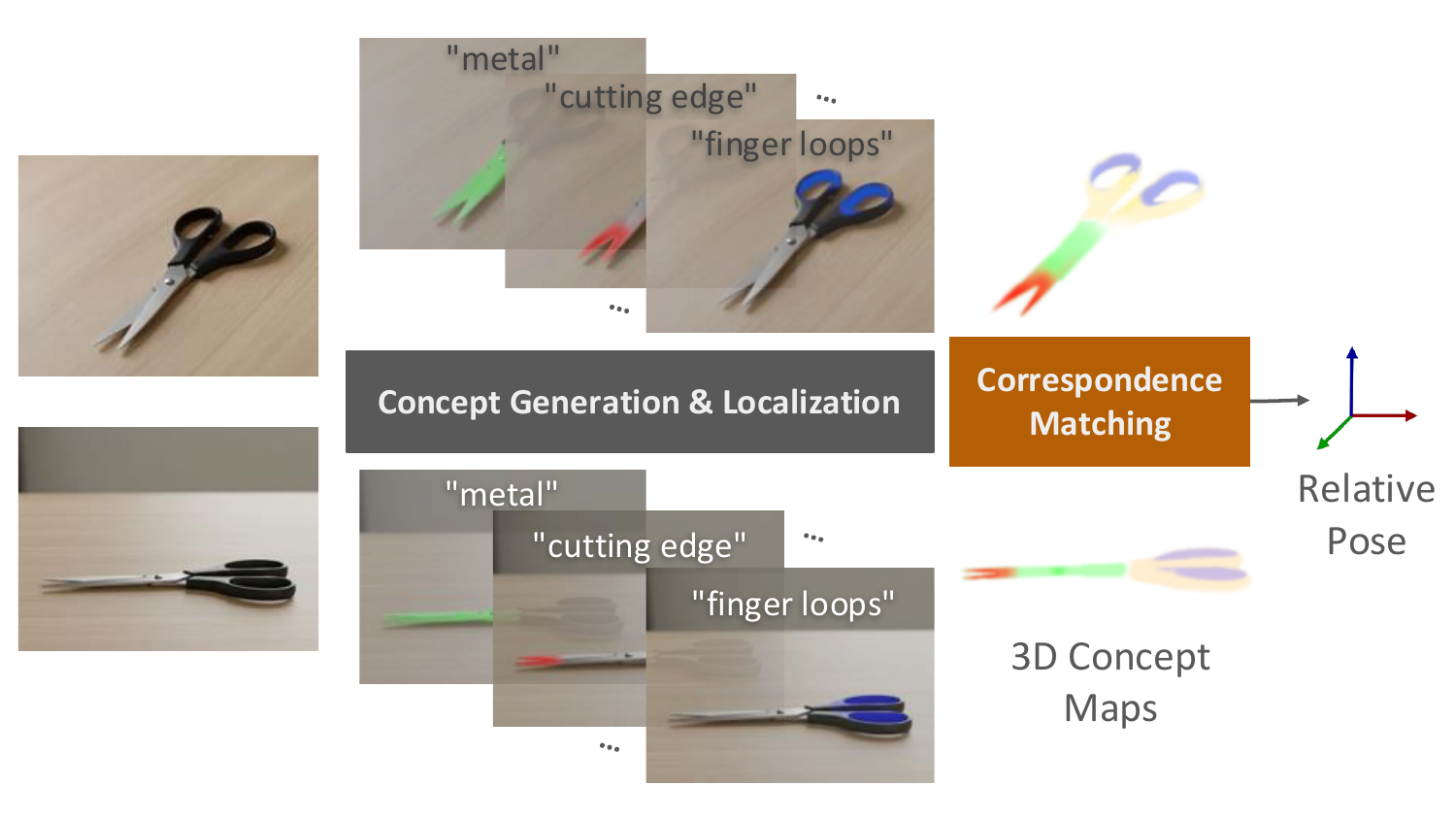}
    \caption{From \textbf{language concepts} to \textbf{6D Pose}: ConceptPose uses language-driven concepts to create 3D concept maps and match them across views for training-free 6D pose estimation.}
    \label{fig:teaser}
\end{figure}

To overcome this rigidity, the community has increasingly focused on category-level~\cite{wang2019,housecat6d2023} and zero shot relative pose estimation~\cite{oryon_corsetti2024,any6d2024,one2any_liu2025,uapose2025,unopose2025}, which aim to handle previously unseen objects. In parallel, a powerful new trend has emerged with the advent of Large-scale Vision Foundation Models (VFMs)~\cite{dino_caron2021,clip_radford2021,siglip2023}. Trained on web-scale data, these models demonstrate unprecedented general-purpose understanding of semantics, geometry, and context—enabling strong zero-shot performance on tasks such as segmentation, detection, and captioning  \cite{2023CLIPDINOiser,2024FindingDino,lidecap}. Naturally, recent 6D pose estimation methods have begun to exploit foundation models. Works such as FoundationPose~\cite{foundationpose2024}, Oryon~\cite{oryon_corsetti2024}, and Horyon~\cite{horyon_corsetti2024} leverage VFM backbones (e.g., DINO~\cite{dino_caron2021}) to extract robust 2D features, which are then used to learn dense correspondences within a trainable pose estimation pipeline. While this strategy achieves state-of-the-art results, it still depends on a crucial training phase: the VFM serves as a frozen feature extractor, but an additional "head" or correspondence network must be trained on top. This limits true generalization and makes adaptation to newer, more powerful VFMs cumbersome.

This work rethinks the approach to object-agnostic pose from a fresh perspective. Consider how humans determine the "pose" of an unseen, unposed object. We begin by noticing distinct characteristics of this novel object. When its pose changes, the mind seeks out those same features from the new viewpoint and establishes cross-view correspondences. This cognitive mechanism makes pose understanding fundamentally object-agnostic. Since language serves as the natural medium for expressing these features, such expressions can be abstracted as \textbf{concepts}. A concept can be anything that describes object characteristics—semantic part descriptions, geometric properties, attributes, or even affordances. Following this reasoning, if machines could also establish spatial relationships through concept recognition, it would naturally unlock a path to derive 6D pose. Thanks to recent advancements in vision language models (VLMs), there now exists a sophisticated way to bridge these two modalities. Just as humans do, by leveraging the power of language as a universal descriptor alongside a pre-trained powerful vision model, existing VLMs can be unlocked to derive precise object 6D pose.


This insight leads to a simple yet remarkably powerful pipeline. We introduce \textbf{ConceptPose}, the first method, to our knowledge, that achieves training-free and model-free zero shot relative pose estimation by leveraging VLMs for language-driven reasoning. We compute the 6DoF relative pose directly by matching semantic concept vectors between two views. This approach requires no dataset-specific training, generalizes naturally to new objects and VLMs, and—as we will demonstrate—achieves state-of-the-art performance, outperforming even trained methods on standard pose estimation benchmarks.

In summary, our main contributions are as follows:
\begin{itemize}
\item We introduce ConceptPose, a novel pipeline for 6D object pose estimation that is completely training-free and model-free (requiring no CAD models).
\item We introduce concept vectors as a general method for object description, combining semantic and geometric understanding through language-driven concept generation and VLM-based spatial localization. 

\item We show that our training-free approach not only generalizes to novel objects but surpasses trained methods by over 62\% in ADD(-S) score, revealing that language-driven semantic reasoning can outperform learned geometric features for pose estimation.
\end{itemize}
\section{Related Works}
\label{sec:related_works}

\begin{figure*}[t]
    \centering
    \includegraphics[width=\textwidth]{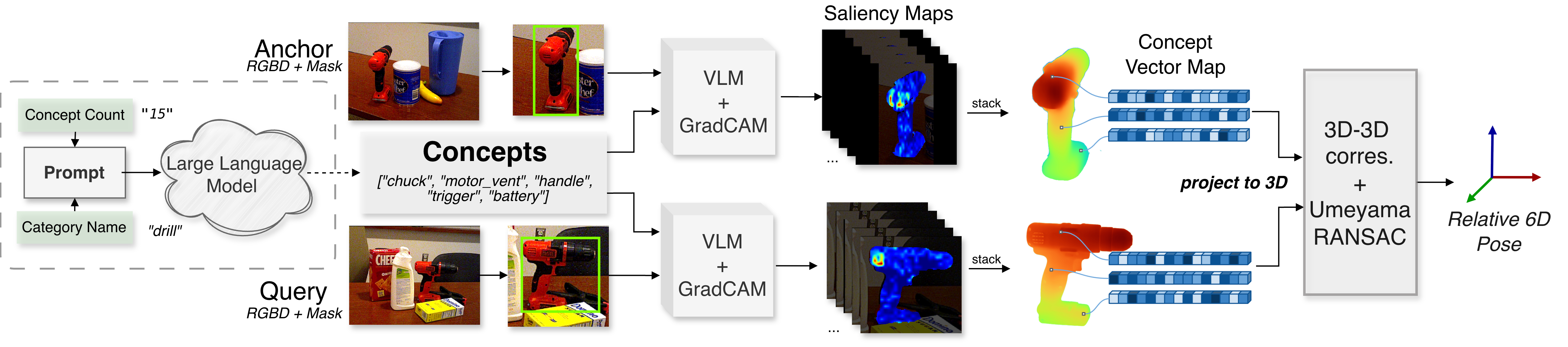}
    \caption{Overview of the ConceptPose pipeline for zero shot relative pose estimation. Given an anchor-query RGB-D pair and category name, we first generate concepts via LLM. These concepts are used to query a VLM to generate dense saliency maps for both frames . The saliency maps are backprojected into 3D and stacked into concept activation vectors, enabling robust semantic correspondence matching for RANSAC-based relative pose estimation.}
    \label{fig:pipeline}
    \vspace{-1em}
\end{figure*}


\noindent\textbf{Model-based Object Pose Estimation} rely on textured CAD models and object-specific training~\cite{hinterstoisser2013,xiang2018,wang2019,hodan2018}. Instance-level approaches~\cite{zebrapose2022,matchu2024,sam6d2024,gigapose2024} learn to estimate poses for specific instances, achieving high accuracy on known objects but failing to generalize to unseen ones. Training-free variants have been proposed for instance-level estimation~\cite{foundpose2024,caraffa2025}, but these still require CAD models at test time, limiting their practical applicability. Category-level methods~\cite{wang2019,housecat6d2023,gspose2024,gcepose2025,Tian2024LearningAC} relax the instance-specific constraint by learning category level representations (e.g., NOCS) that generalize across instances within a category. These methods still require extensive training and are mostly constrained to predefined object categories with available training data. These fundamental limitations motivate model-free, object-agnostic approaches.

\noindent\textbf{Model-free Object-agnostic Pose Estimation} has emerged as a more practical alternative, eliminating the need for CAD models entirely. Multi-view approaches build 3D object representations from reference images or videos. Gen6D~\cite{gen6d2022}, OnePose~\cite{onepose2022}, and OnePose++~\cite{oneposepp2022} use structure-from-motion to reconstruct sparse 3D point clouds and match 2D-3D correspondences. FS6D~\cite{fs6d2022} employs dense RGB-D prototypes with transformers for few-shot pose estimation. FoundationPose~\cite{foundationpose2024} constructs neural object fields from posed RGB-D images, achieving strong pose estimation through differentiable rendering. However, these multi-view methods require multiple reference views for object onboarding, which can be impractical in dynamic environments.

Single-view reference methods have since gained increasing attention. They estimate pose from a single reference image, making them suitable for zero shot or few-shot scenarios. Learning-based methods like NOPE~\cite{nope2024}, One2Any~\cite{one2any_liu2025}, UNOPose~\cite{unopose2025}, and UA-Pose~\cite{uapose2025} train networks to predict pose from single references. H-/Oryon~\cite{oryon_corsetti2024, horyon_corsetti2024} combine DINO~\cite{dino_caron2021} features with text embeddings of only the object name, limiting its further potential. All these approaches require dataset-specific training and cannot easily adapt to new foundation models without retraining.

This motivates training-free alternatives that leverage foundation models directly. Classical methods like SIFT~\cite{lowe2004} and ObjectMatch~\cite{gu2023} perform geometric matching without learning, but lack semantic understanding and struggle with texture-less objects and large viewpoint changes. Recent work has begun exploring training-free pose estimation using VLMs. POPE~\cite{pope2024} employs DINOv2 features for zero-shot pose estimation through direct feature matching between reference and query images, demonstrating strong performance on OnePose benchmarks given non-standard evaluation protocols. While Any6D~\cite{any6d2024} claims training-free, they leverage image-to-3D generation models to reconstruct 3D object models from single reference images, then depends on FoundationPose~\cite{foundationpose2024}—a neural network pretrained on synthetic pose data—for final pose estimation. ConceptPose instead leverages VLM explainability methods to generate dense concept vectors from saliency maps, enabling semantic correspondence without learned pose networks or 3D reconstruction.





\noindent\textbf{Vision Language and Foundation Models } like self-supervised foundation models DINO~\cite{dino_caron2021} and DINOv2~\cite{dinov2_oquab2024} have been adopted in recent pose estimation methods ~\cite{oryon_corsetti2024, horyon_corsetti2024, foundpose2024, pope2024} as feature extractors for their emergent semantic properties.
However, vision-language models (VLMs)—despite their rich semantic understanding from large-scale pretraining~\cite{clip_radford2021, li2022, li2023, zhai2023, siglip2023}—remain underutilized for pose estimation beyond feature extraction.

While language-guided segmentation methods~\cite{ding2023, barsellotti2025talking, ren2024, groundingdino2024} excel at object-level localization through vision-language alignment, they struggle with diverse concepts humans naturally use for pose understanding, like object attributes, affordances and geometric characteristics. VLM explainability methods bridge this gap: techniques like Grad-CAM~\cite{selvaraju2020}, originally designed for model interpretation, provide fine-grained spatial attention for arbitrary text queries that binary masks can't. ConceptPose exploits this capability, repurposing VLM explainability from diagnosis to concept localization. By spatially grounding language-driven concepts, we enable training-free concept-level correspondence for pose estimation without learned matching networks or 3D model, unlocking VLMs for precise spatial reasoning.

\section{Methodology}
\label{sec:methodology}

\subsection{Problem Definition and Notations}

We address relative pose estimation from a single reference view. Given two unposed RGB-D observations of the same object instance from different viewpoints in potentially different scenes---an \emph{anchor} frame $\mathcal{A} = \{I_a, D_a, M_a, K_a\}$ and a \emph{query} frame $\mathcal{Q} = \{I_q, D_q, M_q, K_q\}$---we estimate the 6DoF camera-to-camera transformation $\mathbf{T}_{\text{rel}} = (R_{\text{rel}}, \mathbf{t}_{\text{rel}})$ without dataset-specific training, CAD models, or ground truth object poses during inference. Here $I \in \mathbb{R}^{H \times W \times 3}$ is an RGB image, $D \in \mathbb{R}^{H \times W}$ is a depth map, $M \in \{0,1\}^{H \times W}$ is the ground truth object mask, and $K \in \mathbb{R}^{3 \times 3}$ is the camera intrinsic matrix. Given a category name $c$ (e.g., ``cup''), usually provided by the dataset or specified by the user, ConceptPose automatically generates concepts $\mathcal{L} = \{l_1, \ldots, l_L\}$.

The relative transformation from anchor to query camera frame is defined as $R_{\text{rel}} \in SO(3)$ and $\mathbf{t}_{\text{rel}} \in \mathbb{R}^3$, such that $\mathbf{P}_q = R_{\text{rel}} \cdot \mathbf{P}_a + \mathbf{t}_{\text{rel}}$ for corresponding 3D points $\mathbf{P}_a, \mathbf{P}_q \in \mathbb{R}^3$ in the anchor and query frames respectively. For benchmarking, we convert this to absolute object coordinates by composing with ground truth anchor pose $\mathbf{T}_a^{\text{obj}} = (R_a, \mathbf{t}_a)$: $\mathbf{T}_q^{\text{obj}} = \mathbf{T}_{\text{rel}} \circ \mathbf{T}_a^{\text{obj}}$.

\subsection{Architecture}

Figure~\ref{fig:pipeline} presents an overview of the ConceptPose pipeline. Following, we describe each component in detail.

\noindent\textbf{Category-Level Concept Extraction.} Given a category name $c$ (e.g., ``bottle''), we query a pre-trained general purpose LLM to generate $L$ descriptive concept labels $\mathcal{L} = \{l_1, \ldots, l_L\}$ that characterize the object. Critically, concepts are not limited to semantic parts; they can describe geometry (e.g., ``curved surface'', ``flat base''), affordances (e.g., ``graspable region'', ``pourable opening''), attributes (e.g., ``round'', ``metal''), or any visually localizable property. We use a structured prompt that requests concepts that are: (1) generalizable across different instances, (2) externally visible from at least one viewpoint, and (3) semantically orthogonal to minimize redundancy. This flexible concept formulation enables adaptation to different object properties on the fly. 

In section \ref{sec:ablation-studies}, we provide ablation studies on the number of concepts and different prompt types and additional information on concept generation.

\noindent\textbf{Saliency Map Extraction.} For each concept $l_i \in \mathcal{L}$, we generate a spatial saliency map using a pre-trained VLM. We employ GradCAM~\cite{selvaraju2020} on the vision encoder to compute gradients with respect to the text prompt ``$l_i$.''. We first crop the input RGB image to the object's bounding box and resize to the VLM's input dimension. GradCAM computes class activation maps from the vision transformer's output by weighted summation of gradients and activations, producing saliency maps at the input resolution. We finally resize the saliency map to the object's bounding box size and pad back to the original image dimensions, yielding a $(L, H, W)$ saliency tensor where each channel highlights regions semantically aligned with concept $l_i$. To accelerate inference, we cache text embeddings across all frames. 

\noindent\textbf{Projection and Correspondence.} We backproject the 2D saliency maps to 3D by associating each valid depth pixel with its $(L)$-dimensional concept vector, yielding dense point clouds $\mathcal{P}_a \in \mathbb{R}^{N_a \times 3}$ for the anchor and $\mathcal{P}_q \in \mathbb{R}^{N_q \times 3}$ for the query frame. Before establishing correspondences, we apply two-stage statistical filtering: local outlier removal using k-nearest neighbors, then global outlier removal based on distance to point cloud center. For each 3D point $\mathbf{p} \in \mathcal{P}$, we associate a \textbf{concept vector} $\mathbf{c}(\mathbf{p}) \in \mathbb{R}^L$ extracted from the saliency maps. Each concept vector is normalized via softmax with temperature $\tau$ to form a probability distribution:
\begin{equation}
\mathbf{c}(\mathbf{p}) = \text{softmax}\left(\tfrac{\mathbf{s}(\mathbf{p})}{\tau}\right), \quad c_i(\mathbf{p}) = \tfrac{\exp(s_i(\mathbf{p})/\tau)}{\sum_{j=1}^{L} \exp(s_j(\mathbf{p})/\tau)}
\end{equation}
where $\mathbf{s}(\mathbf{p}) \in \mathbb{R}^L$ denotes the raw saliency values. 

To establish correspondences, we compute a similarity matrix $\mathbf{S} \in \mathbb{R}^{N_q \times N_a}$ using forward KL divergence:
\begin{equation}
S_{ij} = -D_{\text{KL}}(\mathbf{c}(\mathbf{p}_q^i) \| \mathbf{c}(\mathbf{p}_a^j)) = -\sum_{k=1}^{L} c_k(\mathbf{p}_q^i) \log \frac{c_k(\mathbf{p}_q^i)}{c_k(\mathbf{p}_a^j)},
\end{equation}
where $\mathbf{p}_q^i \in \mathcal{P}_q$ and $\mathbf{p}_a^j \in \mathcal{P}_a$. The correspondence for each query point is:
\begin{equation}
\mathbf{p}_a^{*(i)} = \argmax_{\mathbf{p}_a^j \in \mathcal{P}_a} S_{ij} = \argmin_{\mathbf{p}_a^j \in \mathcal{P}_a} D_{\text{KL}}(\mathbf{c}(\mathbf{p}_q^i) \| \mathbf{c}(\mathbf{p}_a^j)),
\end{equation}
yielding a set of putative 3D-3D point pairs for RANSAC-based pose estimation.
Additionally, for computational efficiency, we optionally voxelize the point clouds to produce sparse representations; see Section~\ref{sec:ablation-studies} for details. 

\noindent\textbf{Pose Estimation.} We estimate $\mathbf{T}_{\text{rel}} = (R_{\text{rel}}, \mathbf{t}_{\text{rel}})$ using RANSAC~\cite{fischler1981}-based robust estimation: for each iteration, we sample a minimal set of correspondences and apply Umeyama's~\cite{umeyama1989} closed-form solution to compute a similarity transformation $(R, \mathbf{t})$, then count inliers whose transformed distances fall below a threshold. RANSAC selects the transformation with maximum inliers, effectively optimizing for geometric consistency despite correspondence outliers. Finally, we apply ICP refinement using geometric nearest neighbor matching to locally optimize the alignment. The resulting transformation directly represents the camera-to-camera motion from anchor to query frame.

\section{Experiment and Results}
\label{sec:exp_results}

\begin{figure*}[t]
    \centering
    \includegraphics[width=\textwidth]{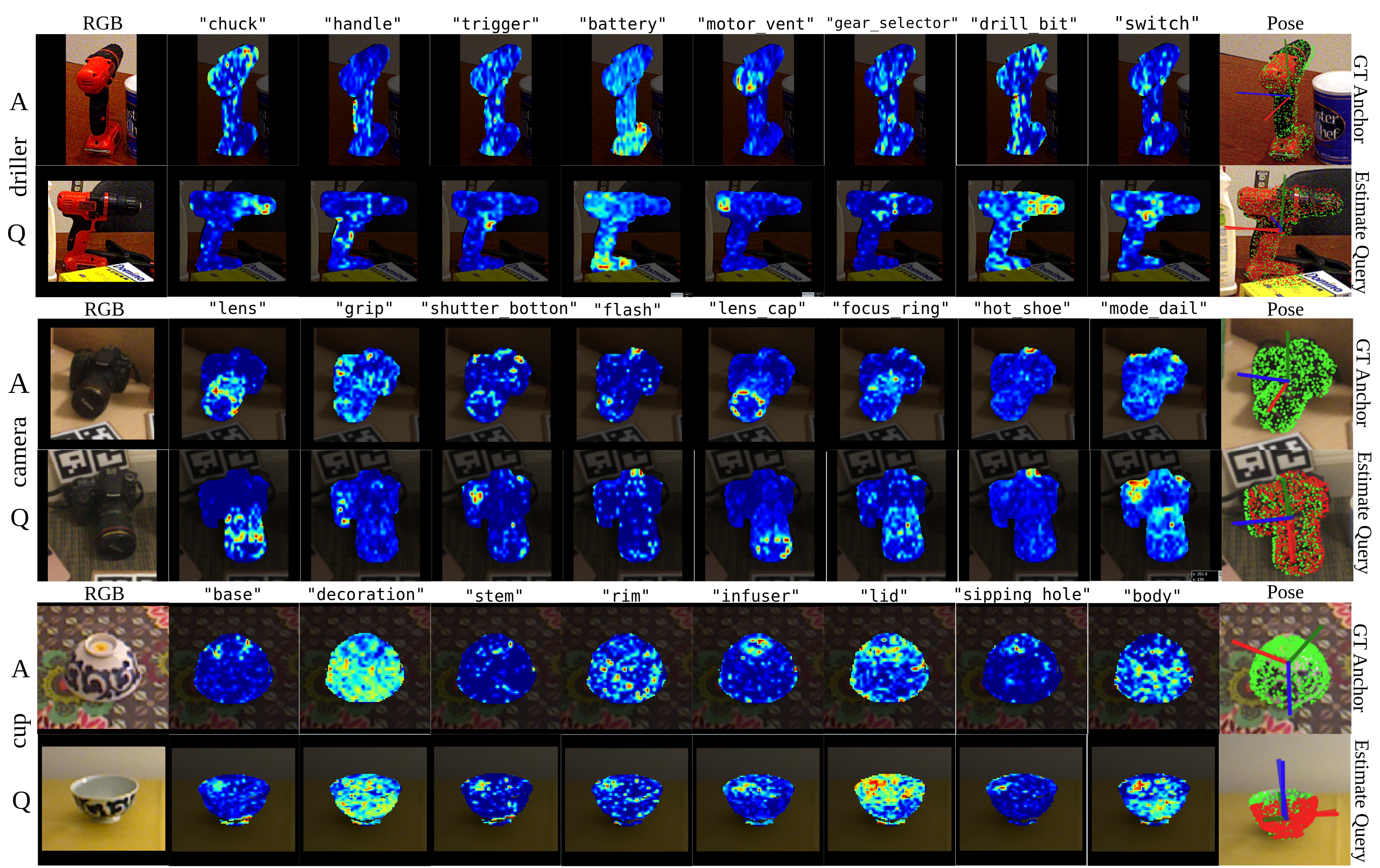}
    \caption{Qualitative results of ConceptPose's zero shot relative pose estimation on REAL275, Toyota-Light, and YCB-Video. For each example, the first column displays the cropped RGB image, followed by eight columns showing distinct semantic concepts extracted for the object category along with their corresponding saliency maps. The final column presents the  \textcolor{green!60!black}{ground truth anchor pose} (top) and the \textcolor{red}{estimated query pose} vs. the \textcolor{green!60!black}{ground truth query pose} (bottom), obtained by applying the estimated relative transformation to the ground truth anchor pose. Notably, concept localization succeeds even on semantically simple symmetric objects with few distinctive parts, such as correctly identifying the base of an upside-down cup.}
    \label{fig:qualitative-results}
    \vspace{-1em}
\end{figure*}

\subsection{Experiment Setup}
\label{sec:experiment-setup}
All experiments are conducted on AMD Ryzen 7 5800X 8-core CPU, and NVIDIA GeForce RTX 4060 Ti (16G), with 48GB of RAM. Our experiments uses PyTorch 2.0+ with CUDA 12.6. We employ SigLIP2-giant-opt-patch16-384 as the VLM and Gemini 2.5 Pro~\cite{gemini2025} for automated concept generation. Notably we are using FP16 precision for all operations to reduce memory usage on consumer-level hardware. We probe SigLIP's vision encoder's \texttt{post\_layernorm} layer with pytorch-gradcam's implementation. Unless otherwise specified, we report all results using our default prompt with $L=20$ for concept label generation, in our experiements, concepts are generated once and they applied to all instances within the category. We use deterministic RANSAC with a fixed random seed (seed=42) across all experiments to ensure reproducibility. For global ourlier removal, we discarding points beyond $\mu + 2.5\sigma$ from the point cloud center. We perform 100,000 RANSAC iterations with a threshold of 0.01m.

\subsection{Datasets}
\label{sec:datasets}
We evaluate ConceptPose on four widely-adopted real-world RGB-D datasets, including REAL275, Toyota-Light, YCB-Video, and LINEMOD. For all datasets, we use ground truth object masks during evaluation to isolate the target object, following standard zero shot relative pose estimation protocols. 

\noindent\textbf{NOCS REAL275}~\cite{wang2019} dataset is a real-world category-level benchmark consisting of 2,754 RGB-D test frames across 6 indoor tabletop scenes. It contains 18 object instances from 6 categories (bottle, bowl, camera, can, laptop, mug), with 3 instances per category.

\noindent\textbf{Toyota-Light (TYOL)}~\cite{bop_2018} is a real-world dataset featuring 21 everyday object classes in realistic tabletop scenarios. The test split contains 21 scenes with varying illumination, clutter, and occlusion conditions Objects include household items such as mugs, plates, remote controls, and magazines.

\noindent\textbf{YCB-Video (YCB-V)}~\cite{xiang2018} is a highly cluttered dataset containing 21 unique objects from the YCB object set. YCB-V is characterized by significant object occlusions, challenging lighting variations, and dense clutter.

\noindent\textbf{LINEMOD (LM)}~\cite{hinterstoisser2013} is a classic 6D pose dataset featuring 15 unique objects including household items and tools. It includes challenging symmetrical objects (e.g., eggbox, glue, bowl).

\subsection{Metrics}
\label{sec:metrics}
To closely follow baseline protocols on relative pose estimation, we consider the following metrics:

\noindent\textbf{ADD-based Metrics.} ADD measures pose accuracy via mean 3D distance between transformed model points. ADD-S uses nearest-neighbor matching for symmetric objects to handle rotational equivalence. ADD(-S) adaptively selects between the two based on symmetry. We report:
\begin{itemize}
 \item \textbf{Threshold-based:} ADD-0.1d recall (ADD, ADD-S, or ADD(-S) $<$ 10\% of object diameter)
 \item \textbf{Curve-based:} ADD-AUC and ADD-S-AUC (area under precision-recall curve up to 0.1m error)
\end{itemize}

\noindent\textbf{BOP Average Recall.} Following the BOP Challenge protocol~\cite{hodan2018}, we compute Average Recall (AR) across three complementary metrics: \textit{VSD} (Visible Surface Discrepancy) evaluates pose quality via depth rendering with occlusion handling, \textit{MSSD} (Maximum Symmetry-Aware Surface Distance) measures worst-case 3D point displacement, and \textit{MSPD} (Maximum Symmetry-Aware Projection Distance) captures 2D reprojection accuracy. For VSD and MSSD, recall is computed across 10 thresholds ranging from 5\% to 50\% of the object diameter. For MSPD, thresholds span 5\% to 50\% of the image dimension. Each metric's AR is the mean recall across its threshold range, and the final BOP Score averages the three AR values. All metrics incorporate symmetry handling to avoid penalizing perceptually equivalent poses.

\noindent\textbf{Additional Metrics.} We report rotation and translation errors, recall at standard thresholds ($5^\circ/2$cm, $10^\circ/5$cm), and 3D Intersection-over-Union (IoU) for volume overlap assessment.

\begin{table*}[t]
    \centering
    \small
    \setlength{\tabcolsep}{5pt}
    \caption{Relative pose estimation performance comparison with recent approaches on REAL275, Toyota-Light, YCB-Video, and LINEMOD. We report the recall of ADD(-S) and BOP AR scores to align with the baseline reports. We report the (\%) improvement over the strongest baseline in green. By default, Average columns show mean performance across all four datasets. The ones marked with $^\dagger$ denotes averages over only REAL275 and TYOL for fair comparison with Any6D and One2Any. TF indicates training-free methods. In Table~\ref{tab:oneshot-voxelization-with-runtime}, we also provide the full metrics of our method with and without voxelization.}
    \resizebox{\textwidth}{!}{%
    \begin{tabular}{l|c|cc|cc|cc|cc|cc}
        \toprule
        \multirow{2}{*}{Method} & \multirow{2}{*}{TF} & \multicolumn{2}{c|}{REAL275} & \multicolumn{2}{c|}{Toyota-Light} & \multicolumn{2}{c|}{YCB-Video} & \multicolumn{2}{c|}{LINEMOD} & \multicolumn{2}{c}{Average} \\
         & & ADD(-S) & BOP AR & ADD(-S) & BOP AR & ADD(-S) & BOP AR & ADD(-S) & BOP AR & ADD(-S) & BOP AR \\
        \midrule
        SIFT~\cite{lowe2004} & \textcolor{gray}{\ding{55}} & 21.6 & 38.8 & 16.5 & 32.4 & 13.9 & 19.3 & 10.8 & 18.7 & 15.7 & 27.3 \\
        ObjectMatch~\cite{gu2023} & \textcolor{gray}{\ding{55}} & 13.4 & 26.0 & 5.4 & 9.8 & 3.7 & 6.0 & 11.1 & 12.2 & 8.4 & 13.5 \\
        Oryon~\cite{oryon_corsetti2024} & \textcolor{gray}{\ding{55}} & 34.9 & 46.5 & 22.9 & 34.1 & 12.8 & 19.4 & 20.4 & 25.3 & 22.8 & 31.3 \\
        Horyon~\cite{horyon_corsetti2024} & \textcolor{gray}{\ding{55}} & 51.6 & \underline{57.9} & 25.1 & 33.0 & \underline{22.6} & \underline{28.6} & \underline{27.6} & \textbf{34.4} & \underline{31.7} & \underline{38.5} \\
        Any6D~\cite{any6d2024} & \textcolor{gray}{\ding{55}} & \underline{53.5} & 51.0 & \underline{32.2} & \underline{43.3} & -- & -- & -- & -- & (42.9$^\dagger$) & (47.2$^\dagger$) \\
        One2Any~\cite{one2any_liu2025} & \textcolor{gray}{\ding{55}} & 41.0 & 54.9 & 34.6 & 42.0 & -- & -- & -- & -- & (37.8$^\dagger$) & (48.5$^\dagger$) \\
        \midrule
        Ours & \textcolor{green!60!black}{\checkmark} & \textbf{71.5} & \textbf{60.4} & \textbf{55.0} & \textbf{51.6} & \textbf{41.2} & \textbf{32.8} & \textbf{38.6} & \underline{31.0} & \textbf{51.6} (\textbf{63.3}$^\dagger$ ) & \textbf{44.0} (\textbf{56.0}$^\dagger$ ) \\
        $\Delta$(\%) & & \textcolor{green!60!black}{+33.6\%} & \textcolor{green!60!black}{+4.3\%} & \textcolor{green!60!black}{+59.0\%} & \textcolor{green!60!black}{+19.2\%} & \textcolor{green!60!black}{+82.3\%} & \textcolor{green!60!black}{+14.7\%} & \textcolor{green!60!black}{+39.9\%} & \textcolor{gray}{-9.9\%} & \textcolor{green!60!black}{+62.8\%} & \textcolor{green!60!black}{+14.3\%} \\
        \bottomrule
    \end{tabular}}
    \label{tab:oneshot-results}
    \vspace{-1em}
\end{table*}

\subsection{Zero-shot Relative Pose Estimation Performance}
\label{sec:pose-estimation-performance}

Following the evaluation protocol proposed by Oryon~\cite{oryon_corsetti2024}, we sample 2000 anchor-query pairs per dataset to assess relative pose estimation performance. For REAL275 and TYOL, we employ Oryon's publicly available fixed splits, which contain 2000 cross-scene only pairs of the same object instance observed from different viewpoints. For YCB-Video and LINEMOD, we generate 2000 pairs via random sampling with a fixed seed filtered to BOP test targets (900 frames for YCB-V, 3000 for LINEMOD). YCB-Video pairs are cross-scene as objects appear in multiple scenes, while LINEMOD pairs are same-scene since each object appears in only one scene. We publish our generated YCB-V and LINEMOD anchor-query pairs using Oryon's format convention for reproducibility.

Figure~\ref{fig:qualitative-results} shows extracted saliency maps and qualitative pose estimation results across the NOCS, YCB-V, and TYOL datasets. We show examples representing both geometrically rich objects with distinct semantic parts and symmetrical objects that pose significant challenges for correspondence-based methods.  Table~\ref{tab:oneshot-results} compares ConceptPose against recent methods across four datasets. Note that Any6D~\cite{any6d2024} and One2Any~\cite{one2any_liu2025} were not evaluated on YCB-Video and LINEMOD, so we omit them from comparison on these benchmarks. For YCB-Video and LINEMOD, different sampling may introduce minor variance, but the performance gaps make comparisons meaningful. 

For the recall of ADD(-S), ConceptPose outperforms all prior work on REAL275 (71.5 vs. 53.5), TYOL (55.0 vs. 32.2), YCB-Video (41.2 vs. 22.6), and LINEMOD (38.6 vs. 27.6), achieving 62.8\% improvements on average over the strongest baselines. For BOP AR, ConceptPose also supasses all baselines on REAL275 (60.4 vs. 57.9), TYOL (51.6 vs. 43.3), YCB-Video (32.8 vs. 28.6), achieving a 14.3\% improvement on average over the second best. It falls short slightly on LINEMOD (31.0 vs. 34.4 of Horyon~\cite{horyon_corsetti2024}) mainly due to the heavy occlusion in that dataset. Since we do not deploy any additional pipeline for occlusion handling whilst still being able to handle occlusion without training to a high degree, we believe this further validates the robustness of our approach. It is a good demonstration that concepts vectors can provide robust correspondence cues across diverse real-world scenarios.

\subsection{Ablation Studies and Analysis}
\label{sec:ablation-studies}

\begin{table*}[t]
    \centering
    \footnotesize
    \setlength{\tabcolsep}{3pt}
    \caption{Ablation study of different prompt types for concepts on the REAL275 dataset. We compare different variants: part-based description (default), geometric descriptions, affordance descriptions, and adjective descriptions. \#L denotes the number of concepts. R indicates whether canonical renderings of category instances are provided to the LLM for additional context.}
    \resizebox{\textwidth}{!}{%
    \begin{tabular}{l|cc|>{\columncolor{gray!15}}c>{\columncolor{gray!15}}c|ccccccc}
        \toprule
        Prompt type & \#L & R & ADD(-S) & BOP AR & ADD & ADD-S & $10^\circ$/5cm & $5^\circ$/2cm & 3DmIoU & 3DIoU50 & 3DIoU75 \\
        \midrule
        default & 15 & \textcolor{gray}{\ding{55}} & \underline{71.5} & \underline{60.4} & \underline{60.6} & 90.8 & \underline{47.2} & \underline{26.0} & \underline{76.4} & \underline{89.4} & \underline{70.1} \\
        geometric & 15 & \textcolor{green!60!black}{\checkmark} & \textbf{71.9} & \textbf{61.1} & \textbf{61.3} & \textbf{91.4} & \textbf{47.8} & \textbf{26.5} & \textbf{76.8} & \textbf{90.0} & \textbf{70.5} \\
        affordance & 15 & \textcolor{green!60!black}{\checkmark} & 68.6 & 58.9 & 57.5 & 90.8 & 44.5 & 23.7 & 75.6 & 88.4 & 67.6 \\
        adjective & 15 & \textcolor{green!60!black}{\checkmark} & 71.2 & 60.6 & 59.8 & \underline{91.3} & 46.3 & 25.2 & 76.3 & \underline{89.4} & 70.1 \\
        \midrule
        default & 20 & \textcolor{gray}{\ding{55}} & \underline{74.0} & \underline{62.4} & \underline{62.4} & \underline{92.5} & \textbf{49.2} & \textbf{27.6} & \underline{77.9} & \underline{92.2} & \underline{71.9} \\
        geo & 20 & \textcolor{green!60!black}{\checkmark} & \textbf{74.6} & \textbf{62.6} & \textbf{63.5} & \textbf{93.8} & \underline{48.2} & \underline{26.8} & \textbf{78.4} & \textbf{93.2} & \textbf{73.0} \\
        \bottomrule
    \end{tabular}}
    \label{tab:concepts-extraction-prompt}
\end{table*}

\noindent\textbf{Concepts Extraction Prompt Ablation.} How prompts are formulated can influence pose estimation performance. We explore four distinct prompting strategies for LLM-based label generation, each emphasizing different aspects of object description. 
\begin{itemize}
    \item The \textit{default} prompt generates generic part names following common object nomenclature (e.g., ``handle'', ``body'', ``spout''). 
    \item The \textit{geometric} prompt produces spatially-grounded descriptions with explicit topology and shape information (e.g., ``top of the handle's curve'', ``central vertical axis of the bottle''). 
    \item The \textit{affordance} prompt combines functional affordances with geometric descriptors, producing rich phrases that capture both form and function (e.g., ``graspable hollow space inside the back handle'', ``pour-able narrow tip of the front spout''). 
    \item The \textit{adjective} prompt adds concise descriptive modifiers to part names, constrained to 2-3 words (e.g., ``curved handle'', ``graspable body''). 
\end{itemize}
The geometric, affordance, and adjective prompts get additionally a rendered image of all objects in each category to provide visual context to the LLM. Table~\ref{tab:concepts-extraction-prompt} presents quantitative results across these prompting strategies on REAL275. Results demonstrate strong robustness across prompting approaches with narrow performance variations. Our default text-only prompt achieves competitive performance (71.5\% ADD(-S), 60.4\% BOP AR), confirming that explicit multimodal input is not strictly necessary for effective concept extraction. Geometric prompts yield marginal but consistent improvements when combined with visual renderings, suggesting that spatially-grounded descriptions provide clearer semantic anchors for VLM-based correspondence matching. In contrast, affordance-based prompts underperform relative to geometric descriptions, indicating that explicit spatial topology is more beneficial than functional affordances for establishing viewpoint-invariant matches. These results confirm that ConceptPose exhibits robustness to prompt design choices while leaving room for modest improvements through prompt engineering and multimodal LLM input.

\medskip

\begin{figure}[t]
    \centering
    \includegraphics[width=\linewidth]{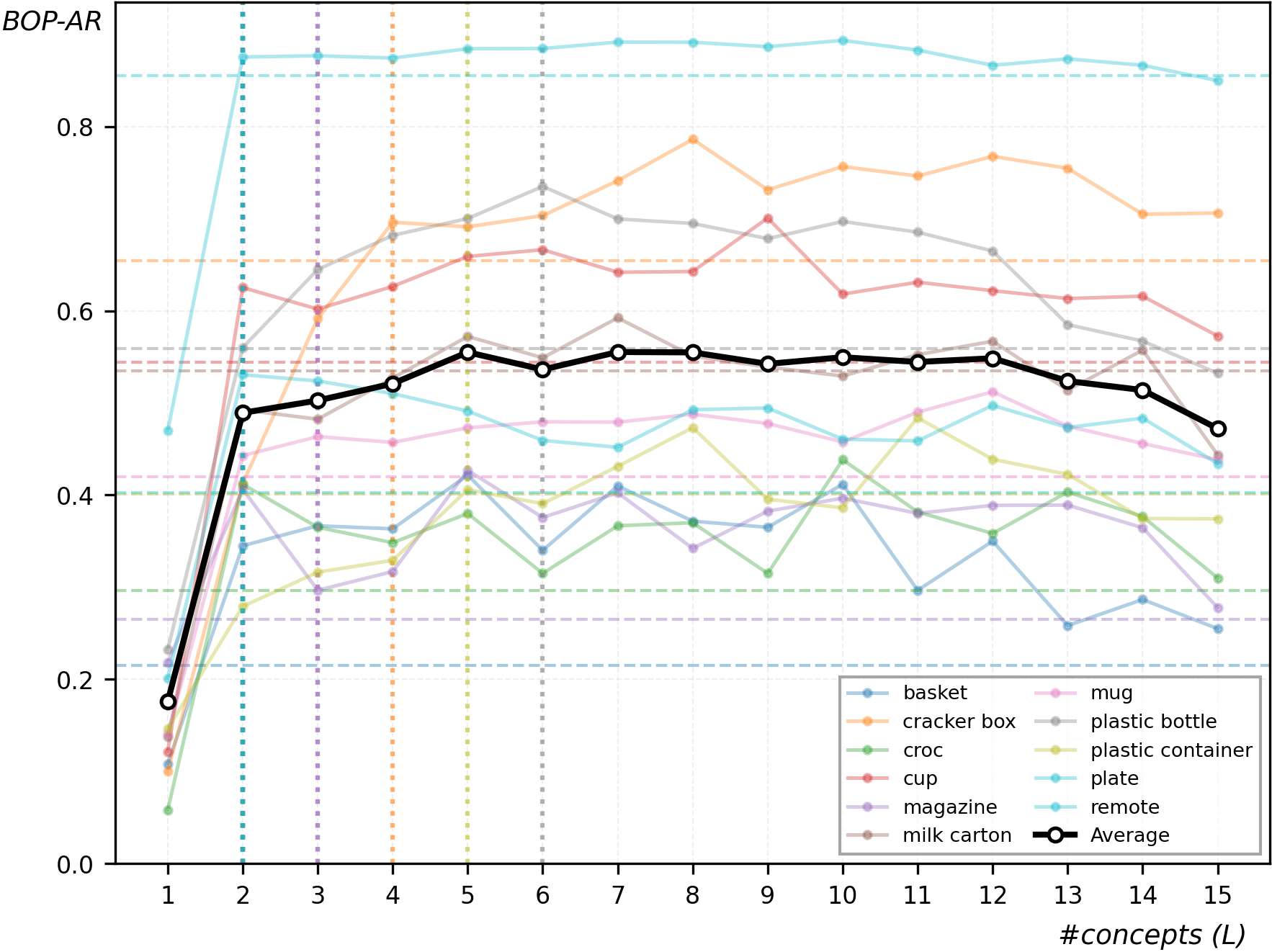}
    \caption{Ablation study on concepts quantity on TYOL dataset.}
    \label{fig:concepts-quantity-study}
    \vspace{-1em}
\end{figure}

\noindent\textbf{Concept Quantity Ablation.}
To determine the optimal number of concepts per object, we conduct greedy oracle-based forward selection analysis on TYOL categories. Starting with an empty label set, the algorithm iteratively adds the single label that maximizes BOP Average Recall (BOP-AR) at each step. Figure~\ref{fig:concepts-quantity-study} shows the resulting performance curves for each category (faded lines), with an average curve for the overall trend. Performance improves rapidly from $L=1$ to $L=4$ concepts, then exhibits diminishing returns with saturation points (vertical dotted lines) occurring around $L=4$-6 for most categories. While the marginal gain becomes negligible, we adopt a uniform $L=15$ across all categories. This is due to the non-deterministic nature of LLM-based concept extraction, we set $L=15$ to ensure maxinum possibility of covering good concepts and avoid hitting the VRAM constraint on consumer level GPUs. This demonstrates our pipeline's generalizability and robustness for in-the-wild deployment without manual hyperparameter tuning.

\noindent\textbf{The Impact of Voxelization.} To improve inference speed, we provide an optional voxelization module that converts dense point clouds $\mathcal{P}_a, \mathcal{P}_q$ into sparse 3D grid representations $\mathcal{V}_a, \mathcal{V}_q$ with $M \ll N$ points. We normalize each point cloud to a unit cube $[-0.5, 0.5]^3$ and discretize it into a $64^3$ voxel grid. Within each occupied voxel, we aggregate the concept activation vectors from all contained points via mean pooling, producing a sparse representation with reduced memory footprint while preserving semantic information. Before pose estimation, we denormalize voxelized points back to camera space using stored scale and centroid (e.g., $\mathcal{V}_a^{\text{cam}} = \mathcal{V}_a^{\text{norm}} \cdot s_a + \mathbf{c}_a$ for the anchor). To fully leverage computational benefits, we reduce max correspondences from 10,000 to 5,000 and RANSAC iterations from 100,000 to 50,000 when voxelization is enabled. 

 Table~\ref{tab:oneshot-voxelization-with-runtime} evaluates the impact of voxelization on pose estimation accuracy across all four datasets. The results demonstrate remarkably minimal performance degradation: ADD(-S) drops by only 0.3-1.8 percentage points, while BOP AR exhibits similarly small drops of 0.3-1.4 points across datasets. Notably, on some individual metrics, voxelization even slightly improves performance (e.g., ADD-S on REAL275: 90.8$\rightarrow$91.3), likely due to the noise-reduction effect of mean pooling within voxels. In the 3rd column of Table~\ref{tab:oneshot-voxelization-with-runtime}, we also provide runtime comparison with and without voxelization. Notably on dataset YCB-V we observe the largest speedup, mainly due to the object is much larger than others, resulting in more projected 3D points. Voxelization normalizes the runtime across datasets, making it more practical to use in real-world applications without sacrificing pose estimation quality. 

\begin{table*}[t]
    \centering
    \small
    \setlength{\tabcolsep}{3pt}
    \caption{Ablation study on the impact of voxelization on relative pose estimation performance and runtime across datasets. Time indicates runtime in seconds per image pair. Voxelization (\textcolor{green!60!black}{\checkmark}) consistently reduces mean runtime from 7.65s to 6.80s (11\% speedup) with mixed effects on performance metrics depending on the dataset.}
    \resizebox{\textwidth}{!}{%
    \begin{tabular}{l|cc|>{\columncolor{gray!15}}c>{\columncolor{gray!15}}c|cccccccccc}
        \toprule
        Dataset & Voxel. & Time & ADD(-S) & BOP AR & ADD & ADD-S & MSSD AR & MSPD AR & VSD AR & $10^\circ$/5cm & $5^\circ$/2cm & 3DmIoU & 3DIoU50 & 3DIoU75 \\
        \midrule
        \multirow{2}{*}{REAL275} & \textcolor{gray}{\ding{55}} & 7.27 & \textbf{71.5} & \textbf{60.4} & \textbf{60.6} & 90.8 & 65.3 & \textbf{69.6} & \textbf{46.2} & \textbf{47.2} & \textbf{26.0} & 76.4 & 89.4 & 69.4 \\
         & \textcolor{green!60!black}{\checkmark} & \textbf{6.87} & 71.2 & 60.1 & 60.4 & \textbf{91.3} & \textbf{65.5} & 69.2 & 45.4 & 46.5 & 24.6 & \textbf{76.7} & \textbf{89.9} & \textbf{70.4} \\
        \midrule
        \multirow{2}{*}{TYOL} & \textcolor{gray}{\ding{55}} & 7.29 & \textbf{55.0} & \textbf{51.6} & \textbf{30.9} & \textbf{85.2} & \textbf{49.1} & \textbf{57.6} & \textbf{48.1} & \textbf{37.6} & \textbf{25.3} & \textbf{70.8} & \textbf{82.4} & \textbf{55.4} \\
         & \textcolor{green!60!black}{\checkmark} & \textbf{6.75} & 53.5 & 50.2 & 29.3 & 83.9 & 47.8 & 56.4 & 46.3 & 35.6 & 22.0 & 69.9 & 81.3 & 53.3 \\
        \midrule
        \multirow{2}{*}{YCB-V} & \textcolor{gray}{\ding{55}} & 8.82 & \textbf{41.2} & \textbf{32.8} & 21.0 & 76.1 & \textbf{37.0} & \textbf{28.2} & \textbf{33.2} & 23.6 & \textbf{11.8} & 59.5 & 65.0 & \textbf{37.7} \\
         & \textcolor{green!60!black}{\checkmark} & \textbf{6.81} & 40.8 & 32.4 & \textbf{22.1} & \textbf{77.8} & 36.7 & 28.1 & 32.4 & \textbf{24.0} & 10.8 & \textbf{60.0} & \textbf{65.5} & 37.5 \\
        \midrule
        \multirow{2}{*}{LINEMOD} & \textcolor{gray}{\ding{55}} & 7.20 & \textbf{38.6} & \textbf{31.0} & \textbf{25.6} & 71.0 & \textbf{31.7} & \textbf{36.1} & \textbf{25.1} & \textbf{25.1} & \textbf{14.0} & 58.5 & \textbf{61.9} & \textbf{32.8} \\
         & \textcolor{green!60!black}{\checkmark} & \textbf{6.75} & 36.8 & 30.3 & 24.8 & \textbf{71.5} & 31.1 & 35.5 & 24.3 & 23.7 & 12.2 & 58.5 & 61.4 & 32.1 \\
        \bottomrule
    \end{tabular}}
    \label{tab:oneshot-voxelization-with-runtime}
    \vspace{-1em}
\end{table*}

\noindent\textbf{Extended Experiment on Few-shot Pose Tracking.} Following FoundationPose~\cite{foundationpose2024} and UA-Pose~\cite{uapose2025}, we present an additional evaluation of few-shot pose tracking on the test split of YCB-Video filtered to BOP test targets. While our pipeline was not specifically designed for this use case, we evaluate it to demonstrate the flexibility of our concept-based approach and its generalizability to related tasks. For each object instance within a scene, we select 2 reference frames using icosphere-based farthest point sampling on the viewing sphere to maximize viewpoint diversity, matching FoundationPose's reference selection strategy. These reference frames are used to build a instance-level concept model via multi-view aggregation of semantic saliency maps. We then evaluate pose estimation on all remaining frames within the same scene, excluding the reference frames from the test set. In total we evaluate 4068 A-Q pairs.
Table~\ref{tab:tracking-results} shows few-shot tracking results on YCB-Video.  ConceptPose achieves 90.1\% ADD-AUC and 95.4\% ADD-S-AUC using 2 static reference frames, outperforming FoundationPose (87.4\%/94.3\%) without training. While UA-Pose achieves higher accuracy (92.8\%/96.5\%) due to online object completion that progressively refines the model during testing, both ConceptPose and FoundationPose use fixed models throughout. Our 2.7-point ADD-AUC gain over FoundationPose demonstrates that concept vectors enable competitive few-shot performance across large viewpoint variations.

\noindent\textbf{Limitations.}
ConceptPose inherits computational characteristics shared with other training-free VLM-based approaches~\cite{foundationpose2024,sam6d2024}. Runtime remains approximately 7 seconds per image pair (6.8 seconds with voxelization), stemming primarily from VLM inference overhead—a limitation shared across methods leveraging dense vision-language models. As VLM architectures advance, we expect proportional runtime improvements. Performance degrades under extreme viewpoint changes for highly asymmetric objects and severe occlusions, challenges common to correspondence-based methods~\cite{foundationpose2024,gen6d2022}. A promising direction is extending ConceptPose to category-level training-free pose estimation requiring no reference view at all—leveraging concept vectors' inherent category-level nature to estimate poses directly from category names, eliminating even the one-shot requirement.

\begin{table}[t]
    \centering
    \small
    \setlength{\tabcolsep}{4pt}
    \caption{Few-shot pose tracking on YCB-V. We report ADD-AUC and ADDS-AUC scores (\%). TF indicates training-free methods. CF indicates completion-free methods (i.e., without online shape completion). Our method achieves comparable performance to state-of-the-art while being both training-free and completion-free.}
    \resizebox{\columnwidth}{!}{%
    \begin{tabular}{l|cc|cc}
        \toprule
        Method & TF & CF & ADD-AUC & ADDS-AUC \\
        \midrule
        FoundationPose~\cite{foundationpose2024} & \textcolor{gray}{\ding{55}} & \textcolor{green!60!black}{\checkmark} & 87.4 & 94.3 \\
        UA-Pose~\cite{uapose2025} & \textcolor{gray}{\ding{55}} & \textcolor{gray}{\ding{55}} & \textbf{92.8} & \textbf{96.5} \\
        \midrule
        Ours & \textcolor{green!60!black}{\checkmark} & \textcolor{green!60!black}{\checkmark} & \underline{90.1} & \underline{95.4} \\
        \bottomrule
    \end{tabular}}
    \label{tab:tracking-results}
    \vspace{-1em}
\end{table}
\section{Conclusion}
\label{sec:conclusion}

We presented ConceptPose, a training-free and model-free approach for 6D object pose estimation that leverages VLM explainability for language-driven semantic reasoning. By querying an LLM to generate concept labels and using VLM saliency maps for spatial localization, we construct dense 3D concept vectors that enable correspondence-based pose estimation without learned matching networks. Our method achieves state-of-the-art results on REAL275~\cite{wang2019} (71.5\% ADD(-S), +33.6\%), TYOL~\cite{bop_2018} (55.0\%, +59.0\%), and YCB-Video~\cite{xiang2018} (41.2\%, +82.3\%), significantly outperforming existing methods including those requiring training—demonstrating that conceptual reasoning can surpass learned geometric features. ConceptPose also demonstrates strong few-shot tracking performance (90.1\% ADD-AUC), surpassing FoundationPose~\cite{foundationpose2024} without training.

Beyond technical achievements, ConceptPose democratizes pose estimation by eliminating the training bottleneck entirely—moving it from servers to and everyday automation. Where traditional methods require extensive training datasets and CAD models, ConceptPose enables agents to adapt on-the-fly to novel objects through natural language concept queries and a single reference view.

We envision ConceptPose paving the way for embodied AI systems that understand objects through human-like conceptual reasoning—identifying graspable surfaces and functional parts through semantic understanding rather than memorized patterns. By making pose estimation queryable, we take a critical step toward embodied agents that interact with the physical world as flexibly and adaptively as humans do.

{
    \small
    \bibliographystyle{ieeenat_fullname}
    \bibliography{main}
}

\clearpage
\setcounter{page}{1}
\maketitlesupplementary


\section{Additional ablations}

\begin{table*}[t]
    \centering
    \small
    \setlength{\tabcolsep}{3pt}
    \caption{Ablation study of Vision-Language Model (VLM) backbones on REAL275 dataset. We compare SigLIP2 variants (giant-384, large-384, base-384) against CLIP ViT-L/14@336px and DINOv3-L/16 with dinotxt. $^\dagger$ indicates our baseline method used throughout the main paper.}
    \resizebox{\textwidth}{!}{%
    \begin{tabular}{l|>{\columncolor{gray!15}}c>{\columncolor{gray!15}}c|cccccccccc}
        \toprule
        Backbone & ADD(-S) & BOP AR & ADD & ADD-S & MSSD AR & MSPD AR & VSD AR & $10^\circ$/5cm & $5^\circ$/2cm & 3DmIoU & 3DIoU50 & 3DIoU75 \\
        \midrule
        SigLIP2-giant-384$^\dagger$ & \textbf{72.0} & \textbf{60.9} & \textbf{60.8} & \textbf{91.2} & \textbf{66.0} & \textbf{70.1} & \textbf{46.6} & \textbf{47.4} & \textbf{26.1} & \textbf{76.8} & \textbf{90.0} & \textbf{69.8} \\
        SigLIP2-large-384 & \underline{67.0} & \underline{56.8} & \underline{56.2} & \underline{89.6} & \underline{61.8} & \underline{65.5} & \underline{43.2} & \underline{43.9} & \underline{22.4} & \underline{74.0} & \underline{86.1} & \underline{64.8} \\
        SigLIP2-base-384 & 63.8 & 54.0 & 53.2 & 88.6 & 59.1 & 62.2 & 40.6 & 40.9 & 24.5 & 72.8 & 84.8 & 61.3 \\
        CLIP ViT-L/14@336px & 54.1 & 46.3 & 43.8 & 84.4 & 50.3 & 53.3 & 35.3 & 32.5 & 13.8 & 67.4 & 76.5 & 52.4 \\
        DINOv3-L/16 + dinotxt & 62.3 & 52.6 & 52.0 & 87.8 & 57.6 & 60.8 & 39.3 & 40.1 & 18.9 & 71.5 & 82.8 & 59.7 \\
        \bottomrule
    \end{tabular}}
    \label{tab:backbone-ablation}
\end{table*}

\subsection{Backbone Ablation}
To demonstrate ConceptPose's adaptability across different vision-language architectures, we evaluate five backbones on REAL275 (Table~\ref{tab:backbone-ablation}): three SigLIP2~\cite{siglip2023} variants (giant-384, large-384, base-384), CLIP ViT-L/14@336px~\cite{clip_radford2021}, and DINOv3-L/16~\cite{simeoni2025dinov3} with the dinotxt text grounding head~\cite{jose2024dinotxt}. Our baseline method is SigLIP2-giant-384. All experiments maintain identical pipeline configurations without voxelization to isolate the impact of backbone architecture.

\noindent\textbf{SigLIP2 and CLIP (GradCAM-based).} For SigLIP2-giant-384 (our baseline), SigLIP2-large-384, and SigLIP2-base-384, we use Hugging Face implementations with GradCAM~\cite{selvaraju2020} applied to the \texttt{post\_layernorm} layer. All SigLIP2 variants use 384$\times$384 input resolution. For CLIP ViT-L/14@336px, we target \texttt{visual.transformer.resblocks[-1].ln\_1} (layer norm of the final transformer block) at its native 336$\times$336 resolution. Both methods compute image-text similarity through normalized dot products, with GradCAM extracting gradient-weighted spatial activations from intermediate transformer features.

\noindent\textbf{DINOv3 + dinotxt (direct patch similarity).} DINOv3, as a vision foundation model (VFM) without native language alignment, requires the dinotxt text grounding head from the official repository's ``Pretrained heads - Zero-shot tasks with dino.txt'' configuration. The dinotxt head produces 2048-dimensional text embeddings partitioned into two complementary 1024-dim subspaces: the first half aligns with the class token (global semantics), while the second half is explicitly trained to align with patch tokens (spatial semantics). Critically, GradCAM cannot be effectively applied to DINOv3 + dinotxt because the text-patch alignment is learned during dinotxt head training and encoded directly in the feature space---there is no gradient path from text queries to spatial activations at inference time, as text features are pre-encoded and frozen. Instead, we follow the official approach of computing direct cosine similarity between image patch features and the patch-aligned text embeddings (second 1024-dim), which leverages the architecture's inherent spatial grounding without requiring backpropagation. DINOv3 uses 224$\times$224 input resolution following its standard preprocessing.

\noindent\textbf{Results.} We evaluate the performance of ConceptPose with different backbones and report the results in Table~\ref{tab:backbone-ablation}. Our baseline SigLIP2-giant-384 achieves the best performance (72.0\% ADD(-S), 60.9\% BOP AR), followed by SigLIP2-large-384 (67.0\%, 56.8\%) and SigLIP2-base-384 (63.8\%, 54.0\%). DINOv3-L/16 + dinotxt reaches competitive performance (62.3\%, 52.6\%) despite using lower input resolution (224 vs. 384) and a fundamentally different saliency extraction method (direct patch similarity vs. GradCAM). CLIP ViT-L/14@336px shows the weakest results (54.1\%, 46.3\%), likely because its contrastive pre-training emphasizes global image-text matching rather than fine-grained spatial correspondence. The 17.9 percentage point gap between SigLIP2-giant-384 and CLIP ViT-L/14@336px (both using GradCAM with comparable resolutions) highlights the importance of pre-training objectives that encourage spatial grounding. These results reveal a clear positive correlation between VLM capacity and ConceptPose performance, demonstrating that our method benefits directly from stronger foundation models while remaining architecture-agnostic. 

\begin{table*}[t]
    \centering
    \small
    \setlength{\tabcolsep}{3pt}
    \caption{Ablation study on the number of concepts ($L$) on REAL275 dataset. We evaluate performance using 3, 5, 10, 15, and 20 concepts per object. All experiments use the same pipeline configuration without voxelization. $^\dagger$ indicates our default baseline configuration.}
    \resizebox{\textwidth}{!}{%
    \begin{tabular}{l|>{\columncolor{gray!15}}c>{\columncolor{gray!15}}c|cccccccccc}
        \toprule
        \#$L$ & ADD(-S) & BOP AR & ADD & ADD-S & MSSD AR & MSPD AR & VSD AR & $10^\circ$/5cm & $5^\circ$/2cm & 3DmIoU & 3DIoU50 & 3DIoU75 \\
        \midrule
        20 & \textbf{74.0} & \textbf{62.4} & \textbf{62.4} & \textbf{92.5} & \textbf{67.5} & \textbf{71.8} & \textbf{47.9} & \textbf{49.2} & \textbf{27.6} & \textbf{77.9} & \textbf{92.2} & \textbf{71.9} \\
        15$^\dagger$ & \underline{72.0} & \underline{60.9} & \underline{60.8} & \underline{91.2} & \underline{66.0} & \underline{70.1} & \underline{46.6} & \underline{47.4} & \underline{26.1} & \underline{76.8} & \underline{90.0} & \underline{69.8} \\
        10 & 70.3 & 59.0 & 58.7 & 89.6 & 63.8 & 67.8 & 45.4 & 45.5 & 24.6 & 75.7 & 88.6 & 68.3 \\
        5 & 68.3 & 57.6 & 57.3 & 88.5 & 62.3 & 66.3 & 44.2 & 44.9 & 24.4 & 75.3 & 87.6 & 67.8 \\
        3 & 64.3 & 54.9 & 52.6 & 88.4 & 59.2 & 63.0 & 42.6 & 42.2 & 23.3 & 74.5 & 87.2 & 64.5 \\
        \bottomrule
    \end{tabular}}
    \label{tab:concept-number-naive-ablation}
\end{table*}

\subsection{Concept Number Ablation}
\begin{figure}[h]
    \centering
    \includegraphics[width=\columnwidth]{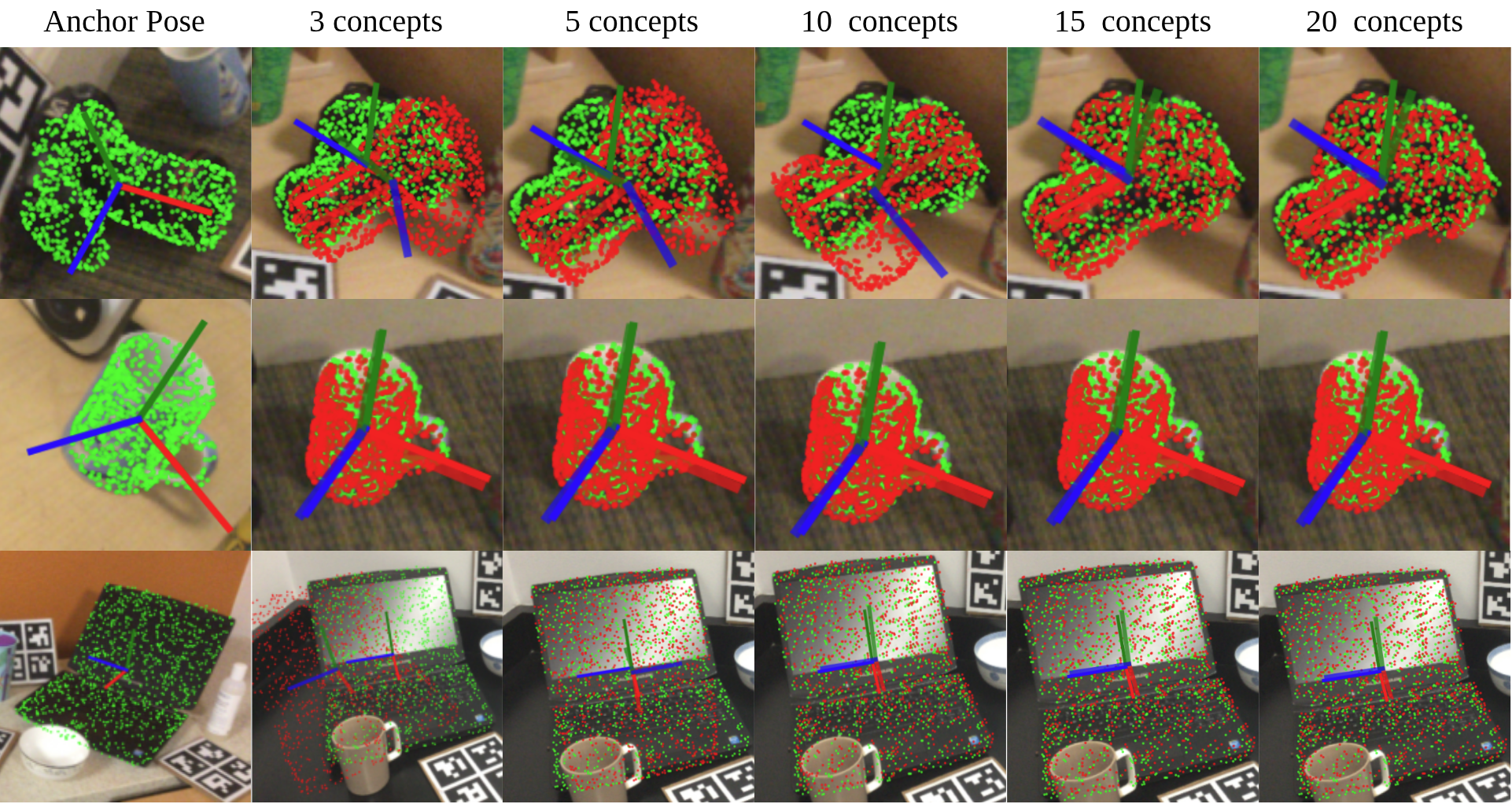}
    \caption{Qualitative visualization of performance changes across different numbers of concepts ($L$) on REAL275 dataset. The first column is the anchor pose, column 2-6 are \textcolor{red}{estimated query poses} with different numbers of concepts ($L$) and \textcolor{green!60!black}{ground truth query pose}.}
    \label{fig:concept-number-ablation}
\end{figure}

To expand on our analysis of the impact of concept quantity in section \textcolor{cvprblue}{4.5}, We present an addidtional evaluation on the number of concepts ($L$). As explain in section \textcolor{cvprblue}{4.1}, we generate $L$=20 concepts per object type by default, however, in this experiment, we truncate the concept list to 3, 5, 10, 15, and 20 concepts per object to study the impact of using less concepts independent of the quality of the concepts. Figure~\ref{fig:concept-number-ablation} shows a qualitative visualization of the performance changes on different object types. All experiments use the same pipeline configuration without voxelization. In Table~\ref{tab:concept-number-naive-ablation}, the results show consistent performance improvements with more concepts, with $L=20$ achieving the best overall performance. However, as our baseline configuration we used $L=15$ in order to balance between performance and computational efficiency.

\subsection{Correspondence Method Ablation}

\begin{table}[ht]
    \centering
    \small
    \setlength{\tabcolsep}{10pt}
    \caption{Ablation study of correspondence methods for matching concept distributions between query and anchor point clouds on REAL275 dataset. We compare five different divergence and similarity measures. All methods use the same pipeline configuration.}
    \label{tab:correspondence-ablation}
    \begin{tabular}{l|cc}
        \toprule
        \textbf{Method} & \textbf{ADD(-S)} & \textbf{BOP AR} \\
        \midrule
        Bidirectional KL & 72.0 & \textbf{0.6107} \\
        KL Divergence & 72.0 & 0.6085 \\
        Reverse KL & 72.0 & 0.6072 \\
        Asymmetric & 72.1 & 0.6053 \\
        Cosine & 72.1 & 0.6042 \\
        \bottomrule
    \end{tabular}
\end{table}

To evaluate the impact of different correspondence methods for matching concept vectors between query and anchor, we compare five divergence and similarity measures on REAL275 (Table~\ref{tab:correspondence-ablation}). All methods achieve similar performance, with ADD(-S) 10cm ranging from 72.0\% to 72.1\% and BOP scores from 0.6042 to 0.6107. Bidirectional KL divergence achieves the best BOP score (0.6107), while Asymmetric and Cosine tie for the highest ADD(-S) at 72.1\%. The minimal performance variation suggests that ConceptPose's effectiveness is robust to the choice of correspondence method, as all tested measures effectively capture the semantic similarity encoded in concept vectors.

\section{Performance Analysis by Object}

\begin{table}[ht]
    \centering
    \footnotesize
    \setlength{\tabcolsep}{4pt}
    \caption{Concepts used for REAL275 evaluation.}
    \begin{tabularx}{\columnwidth}{l|c|X}
        \toprule
        Cat. & \#$L$ & Concepts \\
        \midrule
        bottle & 20 & cap, lid, spout, nozzle, pump, trigger, body, base, neck, shoulder, label, handle, grip, threads, tamper\_evident\_ring, overcap, sleeve, collar, punt, carry\_loop \\
        \midrule
        bowl & 20 & body, rim, bottom, foot, handle, lid, spout, ear, pedestal, inner\_surface, outer\_surface, rib, embossment, medallion, divider, perforation, knob, groove, notch, marking \\
        \midrule
        camera & 20 & lens, camera\_body, viewfinder, display\_screen, grip, shutter\_button, mode\_dial, control\_dial, flash, hot\_shoe, lens\_cap, lens\_hood, eyecup, focus\_ring, zoom\_ring, battery\_door, memory\_card\_door, strap\_lug, lens\_mount, directional\_pad \\
        \midrule
        can & 20 & body, lid, base, rim, pull\_tab, rivet, score\_line, shoulder, cap, nozzle, logo, barcode, nutrition\_facts, ingredients\_list, ridges, seam, net\_content\_statement, warning\_label, recycling\_symbol, product\_image \\
        \midrule
        laptop & 20 & screen, keyboard, touchpad, webcam, hinge, lid, bottom\_case, port, vent, speaker\_grille, logo, bezel, rubber\_foot, power\_button, keyboard\_deck, indicator\_light, fingerprint\_sensor, screw, touchpad\_button, security\_slot \\
        \midrule
        mug & 20 & body, handle, rim, base, interior, lid, sleeve, logo, decorative\_element, handle\_aperture, foot\_ring, slider, sipper\_hole, gasket, thumb\_rest, spout, infuser, carrying\_loop, vent\_hole, spout\_cover \\
        \bottomrule
    \end{tabularx}
    \label{tab:concept-labels-real275}
\end{table}

\begin{table*}[t]
    \centering
    \small
    \setlength{\tabcolsep}{2.5pt}
    \caption{Performance breakdown by object type on REAL275 dataset. We report results across all six object types (bottle, bowl, camera, can, laptop, mug) with the number of anchor-query pairs evaluated for each. }
    \resizebox{\textwidth}{!}{%
    \begin{tabular}{l|c|>{\columncolor{gray!15}}c>{\columncolor{gray!15}}c|ccc|ccccccc}
        \toprule
        Obj. Type & \#Pairs & ADD(-S) & BOP AR & VSD AR & MSSD AR & MSPD AR & ADD-S & ADD & $10^\circ$/5cm & $5^\circ$/2cm & 3DmIoU & IoU@50 & IoU@75 \\
        \midrule
        bottle & 13 & 100.0 & 68.7 & 68.5 & 90.0 & 47.7 & 100.0 & 0.0 & 0.0 & 0.0 & 84.9 & 100.0 & 100.0 \\
        bowl & 150 & 82.7 & 45.6 & 31.8 & 85.2 & 19.9 & 82.7 & 5.3 & 2.0 & 0.0 & 72.9 & 93.3 & 54.0 \\
        camera & 482 & 68.9 & 72.5 & 45.0 & 96.6 & 76.0 & 95.9 & 68.9 & 32.2 & 1.7 & 76.6 & 93.8 & 67.4 \\
        can & 144 & 85.4 & 72.4 & 48.5 & 97.2 & 71.5 & 85.4 & 23.6 & 6.2 & 0.0 & 74.8 & 85.4 & 66.7 \\
        laptop & 651 & 76.8 & 65.0 & 45.7 & 81.9 & 67.5 & 83.3 & 76.8 & 76.3 & 60.7 & 75.4 & 78.8 & 73.7 \\
        mug & 560 & 60.4 & 75.6 & 49.4 & 97.4 & 80.0 & 98.4 & 60.4 & 49.8 & 20.7 & 78.6 & 97.7 & 70.0 \\
        \bottomrule
    \end{tabular}}
    \label{tab:per-category-performance}
\end{table*}

We use the following default prompt to generate concepts. For practical efficiency, we generate one shared concept vocabulary per object type ($L=20$ concepts), you can find the generated concepts in Table~\ref{tab:concept-labels-real275}.

\vspace{0.3em}
\begin{minipage}{0.96\columnwidth}
\small\textit{``Give me \{num\_labels\} labels that describe different concepts for a \{object\_label\}. I will be using these labels for localization, please make sure they are generalizable to different instances within the same category \{object\_label\}, semantically orthogonal to each other, and must be visible from at least one external viewpoint. Also, please use the most common names and do not use positional descriptions. Please just give me all the labels as a python list, no additional explanations please.''}
\end{minipage}%
\vspace{0.5em}

Table~\ref{tab:per-category-performance} shows detailed performance metrics for each object type on REAL275 using our baseline configuration (SigLIP2-giant-384 with $L=15$ concepts). Performance varies significantly across object types, revealing several patterns. Laptop achieves the strongest strict pose accuracy ($5^\circ$/2cm: 60.7\%), likely due to its distinctive geometric features and rich visual texture (keyboard, ports, logos) that provide reliable concept localization. Mug shows the highest BOP AR (75.6\%) despite moderate ADD(-S) (60.4\%), suggesting the method produces visually plausible poses even when metric accuracy is lower. Bowl demonstrates a large gap between ADD-S (82.7\%) and ADD (5.3\%), reflecting the expected behavior for highly symmetric objects where multiple orientations are geometrically valid. Camera and can both achieve strong BOP performance (72.5\%, 72.4\%), indicating robust pose estimation on objects with moderate geometric complexity. Bottle's small sample size (13 pairs) limits statistical significance despite perfect ADD(-S) recall.

\section{Continuous Concept Vectors vs.\ Binary Masks}

Our GradCAM-based saliency provides continuous activations, as opposed to binary masks produced by segmentation-based approaches. To validate the advantage of soft activations, we create binary concept vector masks by thresholding at 0.5 and evaluate on REAL275. As shown in Table~\ref{tab:continuous-vs-binary}, both ADD(-S) (71.5\% $\rightarrow$ 30.5\%) and BOP AR (60.4\% $\rightarrow$ 28.2\%) decrease strongly, indicating that continuous concept vectors are important for establishing accurate correspondences.

\begin{table}[h]
    \centering
    \footnotesize
    \setlength{\tabcolsep}{4pt}
    \caption{Continuous concept vectors vs.\ binary masks (thresholded at 0.5) on REAL275.}
    \vspace{-5pt}
    \resizebox{\columnwidth}{!}{%
    \begin{tabular}{l|>{\columncolor{gray!15}}c>{\columncolor{gray!15}}c|cccc}
        \toprule
         & ADD(-S) & BOP AR & ADD & ADD-S & $10^\circ$/5cm & $5^\circ$/2cm \\
        \midrule
        Concept vectors    & \textbf{71.5} & \textbf{60.4} & \textbf{60.6} & \textbf{90.8} & \textbf{47.2} & \textbf{26.0} \\
        Binary masks (thr.=0.5) & 30.5 & 28.2 & 25.2 & 62.7 & 17.7 & 6.1 \\
        \bottomrule
    \end{tabular}}
    \label{tab:continuous-vs-binary}
\end{table}

\section{Effect of Concept Quantity on Correspondence Quality}

\begin{figure*}[t]
    \centering
    \includegraphics[width=\textwidth]{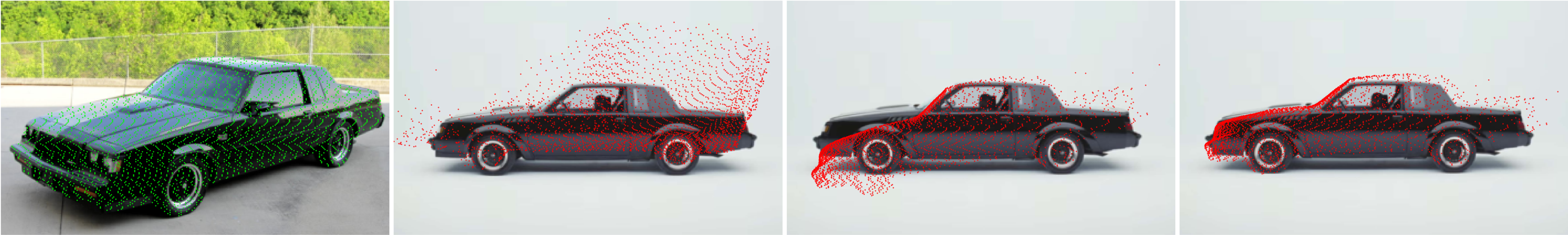}
    \caption{Pose estimation with increasing number of concepts (3, 10, 15) on a car object.}
    \label{fig:concept-quantity-corresp}
\end{figure*}

Spatially extended and repeated parts are a strength of our method: a single spatially localized concept (e.g., car door) can anchor the object, but cannot resolve its scale or orientation. In contrast, a repeated concept (e.g., wheel) helps resolve the object's location, scale, and one additional DoF of orientation. This is analogous to a 2D line segment resolving 4 DoF while a point resolves only 2 DoF. Figure~\ref{fig:concept-quantity-corresp} shows this behavior with increasing number of concepts (3, 10, 15) on a car object.

\section{Occlusion Robustness Analysis}

\begin{figure}[h]
    \centering
    \includegraphics[width=\linewidth]{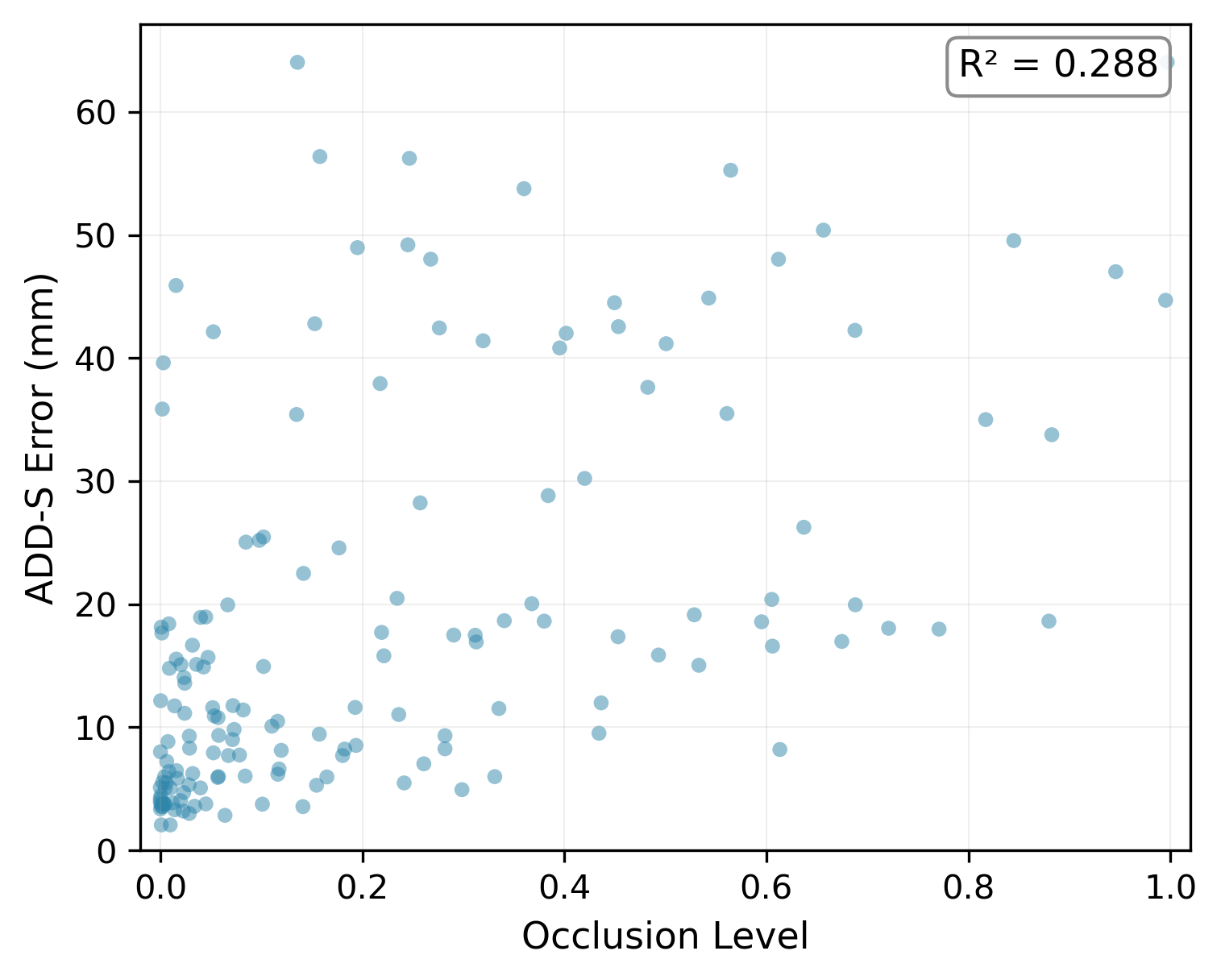}
    \caption{Correlation between occlusion ratio and ADD-S error on LINEMOD-Occlusion. Each subplot shows one object with 20 randomly sampled anchor-query pairs. Regression analysis shows $R^2 < 0.3$, indicating no linear correlation between occlusion and error.}
    \label{fig:occlusion-correlation}
\end{figure}

We present a correlation analysis between the amount of occlusion and pose error (ADD-S) on all objects in LINEMOD-Occlusion (LM-O), sampling 20 anchor-query pairs randomly for each object. Figure~\ref{fig:occlusion-correlation} shows the results. Regression analysis shows $R^2 < 0.3$, indicating no linear correlation between occlusion and error. ConceptPose handles occlusion through several design choices: we crop the input image to the object's bounding box before VLM inference to alleviate potential contamination due to the patch nature of ViTs, backproject only pixels within the object mask to 3D to reduce background noise, and utilize RANSAC's robustness against partial observations by finding the largest geometrically consistent subset among visible points.

\section{Cross-Instance Concept Generalization}

\begin{figure}[h]
    \centering
    \includegraphics[width=0.85\linewidth]{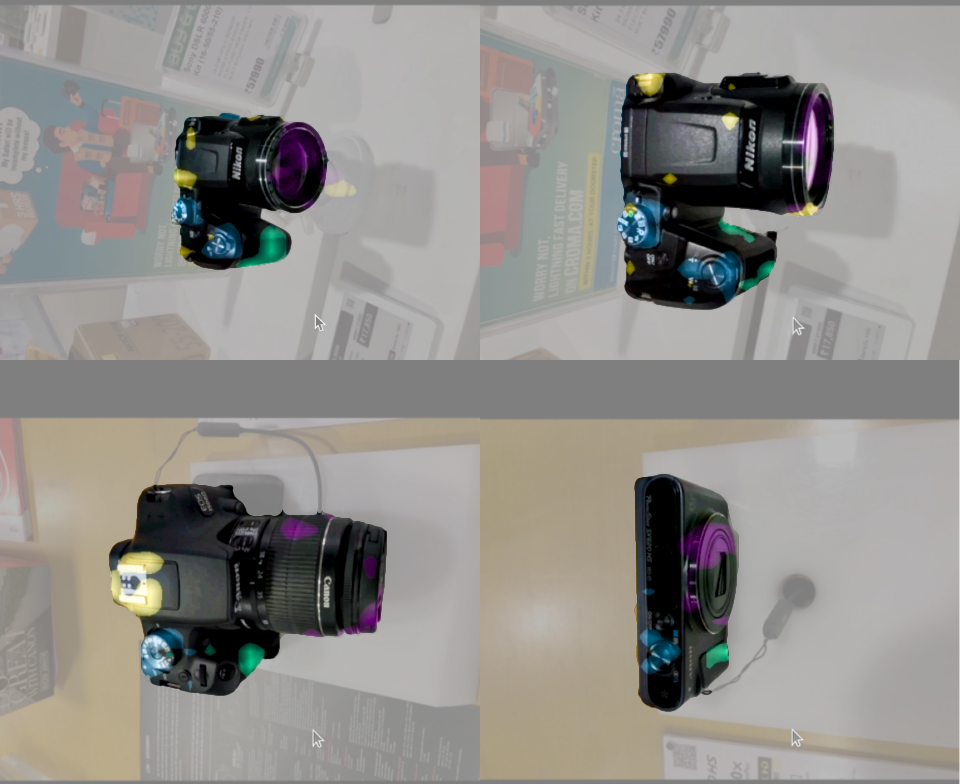}
    \caption{Cross-instance concept activation on different camera instances. The same category-level concepts [\textcolor[rgb]{0.5,0,0.5}{lens}, \textcolor{green}{grip}, \textcolor{blue}{controls}, \textcolor{yellow}{hot\_shoe}] are consistently localized across diverse instances.}
    \label{fig:cross-instance-supp}
\end{figure}

In our task, concepts are inherently shared across instances within a category---not per instance. Figure~\ref{fig:cross-instance-supp} visualizes the same concepts (\textcolor[rgb]{0.5,0,0.5}{lens}, \textcolor{green}{grip}, \textcolor{blue}{controls}, \textcolor{yellow}{hot\_shoe}) activated across different camera instances, demonstrating our training-free approach's ability to generalize across instances, unlike learned geometric descriptors.

\section{Extended Qualitative Results}

\begin{figure*}[p]
    \centering
    \includegraphics[angle=270,width=0.88\textwidth,keepaspectratio]{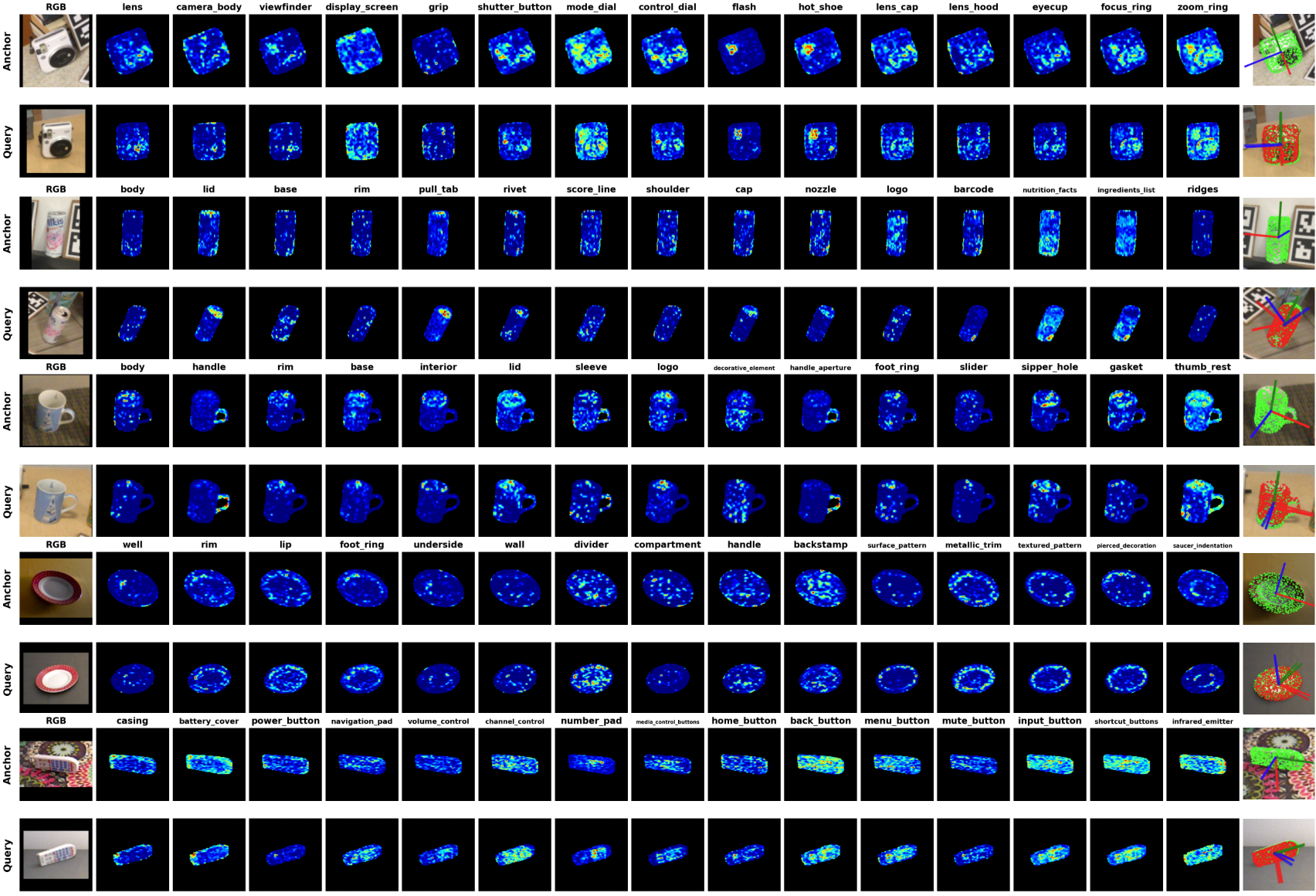}
    \caption{Qualitative results (1/2), \textcolor{red}{estimated query pose}, \textcolor{green!60!black}{ground truth anchor/query pose}.}
    \label{fig:qualitative-1}
\end{figure*}

\begin{figure*}[p]
    \centering
    \includegraphics[angle=270,width=0.88\textwidth,keepaspectratio]{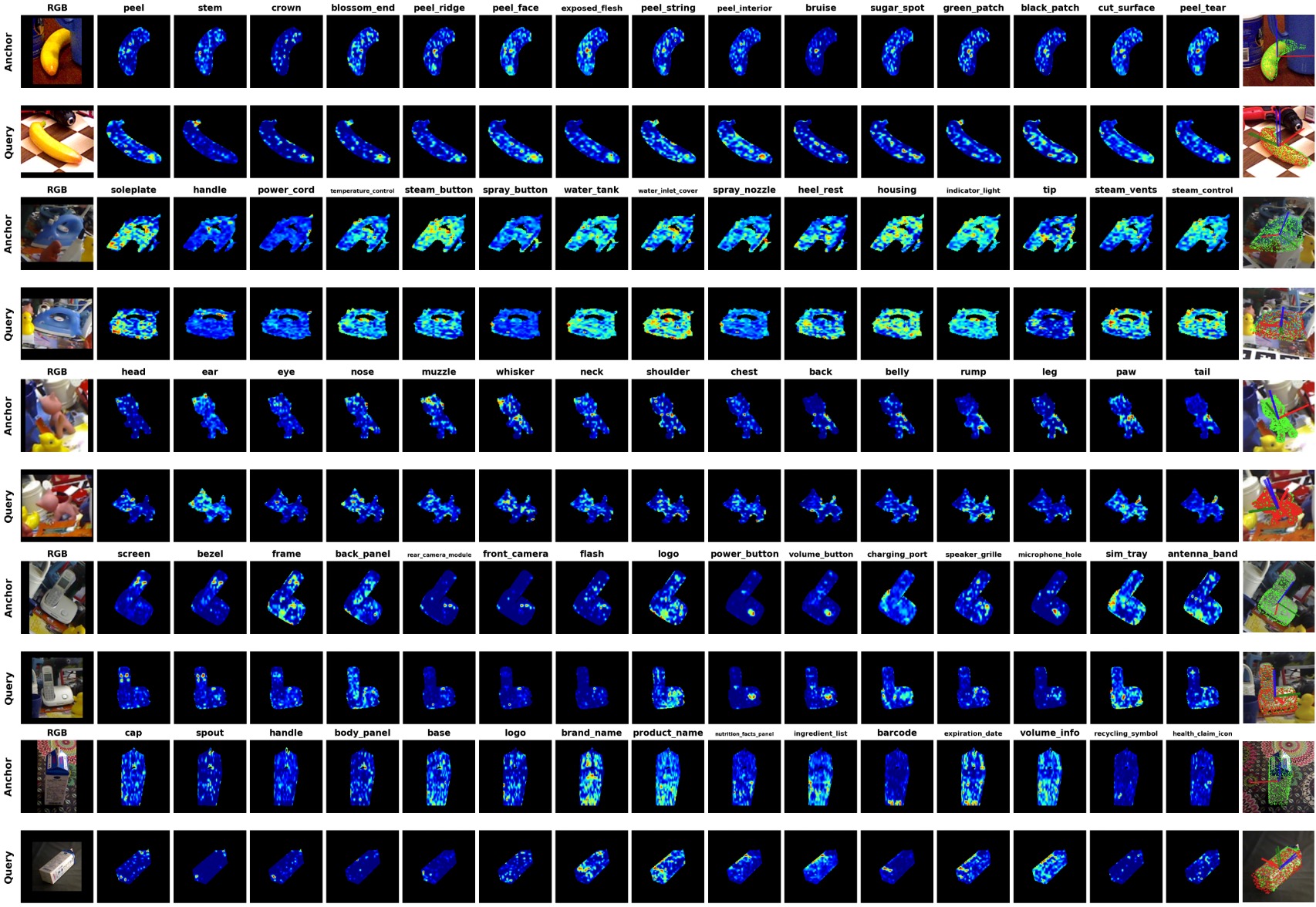}
    \caption{Qualitative results (2/2), \textcolor{red}{estimated query pose}, \textcolor{green!60!black}{ground truth anchor/query pose}.}
    \label{fig:qualitative-2}
\end{figure*}

Figures~\ref{fig:qualitative-1} and \ref{fig:qualitative-2} present 10 extended qualitative visualizations of ConceptPose across all four evaluation datasets (REAL275, Toyota-Light, YCB-Video, LINEMOD), with saliency maps and estimated poses. It demonstrates ConceptPose's ability to generate semantically meaningful concept activations and accurate pose estimates across diverse object categories and imaging conditions.

\end{document}